\documentclass[letterpaper, 10 pt, conference]{ieeeconf}  

\IEEEoverridecommandlockouts                              

\usepackage{amsmath} 
\usepackage{amssymb}  

\usepackage{algorithm}
\usepackage[noend]{algpseudocode}

\usepackage{bm}
\DeclareMathOperator*{\argmax}{arg\,max}
 
\newcommand{\minus}{\scalebox{0.75}[1.0]{$-$}}
\DeclareMathOperator{\sgn}{sgn}

\usepackage{booktabs}

\usepackage[pdftex]{graphicx}
\usepackage{subcaption}

\usepackage{hyperref}

\usepackage[stable]{footmisc}

\usepackage{balance}

\title{\LARGE \bf
Tactical Decision-Making in Autonomous Driving by\\Reinforcement Learning with Uncertainty Estimation}

\author{Carl-Johan Hoel$^{*\dagger}$, Krister Wolff$^{*}$, and Leo Laine$^{*\dagger}$
\thanks{This work was partially supported by the Wallenberg Artificial Intelligence, Autonomous Systems and Software Program (WASP), funded by Knut and Alice Wallenberg Foundation, and partially by Vinnova FFI.}
\thanks{$^{*}$Carl-Johan Hoel, Krister Wolff, and Leo Laine are with Chalmers University of Technology, Gothenburg, Sweden
        {\tt\small \{carl-johan.hoel, krister.wolff, leo.laine\}@chalmers.se}}%
\thanks{$^{\dagger}$Carl-Johan Hoel and Leo Laine are with Volvo Group, Gothenburg, Sweden
        {\tt\small \{carl-johan.hoel, leo.laine\}@volvo.com}}%
}

\begin{document}

\maketitle
\thispagestyle{empty}
\pagestyle{empty}


\begin{abstract}

Reinforcement learning (RL) can be used to create a tactical decision-making agent for autonomous driving. However, previous approaches only output decisions and do not provide information about the agent's confidence in the recommended actions. 
This paper investigates how a Bayesian RL technique, based on an ensemble of neural networks with additional randomized prior functions (RPF), can be used to estimate the uncertainty of decisions in autonomous driving.
A method for classifying whether or not an action should be considered safe is also introduced.
The performance of the ensemble RPF method is evaluated by training an agent on a highway driving scenario. It is shown that the trained agent can estimate the uncertainty of its decisions and indicate an unacceptable level when the agent faces a situation that is far from the training distribution.
Furthermore, within the training distribution, the ensemble RPF agent outperforms a standard Deep Q-Network agent.
In this study, the estimated uncertainty is used to choose safe actions in unknown situations. However, the uncertainty information could also be used to identify situations that should be added to the training process.

\end{abstract}



\section{Introduction}

Autonomous driving has the potential to benefit society in many ways, such as increase the productivity and improve the energy efficiency of autonomous vehicles, and to reduce the number of accidents~\cite{Fagnant2015}. 
The decision-making task of an autonomous vehicle is challenging, since the system must handle a diverse set of environments, interact with other traffic participants, and consider uncertainty in the sensor information.
To manually predict all situations that can occur and code a suitable behavior is infeasible. Therefore, it is compelling to consider methods that are based on machine learning to train a decision-making agent. 
A desired property of such an agent is that it should not just output a recommended decision, but also estimate the uncertainty of the given decision.
This paper investigates a way to create a tactical\footnote{The decision-making task of an autonomous vehicle is commonly divided into strategic, tactical, and operational decision-making, also called navigation, guidance and stabilization~\cite{Michon1985},~\cite{Ulbrich2017}. In short, tactical decisions refer to high level, often discrete, decisions, such as when to change lanes on a highway.} decision-making agent that is also aware of its limitations, for autonomous driving.

Traditional decision-making methods, which are based on predefined rules and implemented as hand-crafted state machines, were successful during the DARPA Urban Challenge~\cite{darpaCMU},~\cite{darpaStanford},~\cite{darpaVirginia}. Other classical methods treat the decision-making task as a motion planning problem~\cite{Werling2010},~\cite{Nilsson2015},~\cite{Damerow2015}. Although these methods are successful in many cases, one drawback is that they are designed for specific driving situations, which makes it hard to scale them to the complexity of real-world driving.

Reinforcement learning (RL) 
techniques have been successful in various domains during the last decade~\cite{Mnih2015}, \cite{Lillicrap2015},~\cite{Silver2018AlphaZero}.
Compared to non-learning based methods, RL approaches are general and could potentially scale to all driving situations.
RL has previously been applied to decision-making for autonomous driving in simulated environments, for example Deep Q-Network (DQN) approaches for highway driving~\cite{Wang2018},~\cite{Hoel2018} and intersections~\cite{Tram2018},~\cite{Isele2018}, 
policy gradient techniques for merging situations \cite{Shalev2016},
or combining Monte Carlo tree search and RL \cite{8911507}.
A few studies have also trained decision-making agents in a simulated environment and then deployed them in a real vehicle~\cite{Pan2017},~\cite{Bansal2018}, or for a limited case, trained an agent directly in a real vehicle~\cite{Kendall2019}.

The agents that were trained by RL in previous works can naturally only be expected to output rational decisions in situations that are close to the training distribution. However, a fundamental problem with these methods is that no matter what situation the agents are facing, they will always output a decision, with no information on the uncertainty of the decision or if the agent has experienced anything similar during its training.
If, for example, an agent that was trained for a one-way highway scenario would be deployed in a scenario with oncoming traffic, it would still output a decision, without any warning.
A more subtle difference could be if the agent has been trained for nominal highway driving, and suddenly faces a speeding driver or an accident that creates standstill traffic.
The importance of estimating the uncertainty of decisions in autonomous driving is further emphasized by McAllister et al.~\cite{McAllister2017}.

A common way to model uncertainty is through Bayesian probability theory \cite{Kochenderfer2015}. Bayesian deep learning has previously been used in the autonomous driving field for, e.g., image segmentation~\cite{Kendall2017} and end-to-end learning~\cite{Michelmore2018}. Early work on applying Bayesian approaches to RL, for balancing the exploration vs. exploitation dilemma, was introduced by Dearden et al.~\cite{Dearden1998}. More recent studies have extended this approach to deep RL, by using an ensemble of neural networks with randomized prior functions~\cite{Osband2018}.

In contrast to the related work, this paper investigates an RL method for tactical decision-making in autonomous driving that can estimate the uncertainty of its decision, based on the work by Osband et al.~\cite{Osband2018} (Sect.~\ref{sec:approach}). A criterion for when the agent is considered confident enough about its decisions is introduced (Sect.~\ref{sec:safeActions}). A decision-making agent is trained in a one-way highway driving scenario (Sect.~\ref{sec:experiments}), and the results show that it outperforms both a common heuristic method and a standard DQN method (Sect.~\ref{sec:results_training_scenarios}).
This study shows that the presented method can estimate the uncertainty of the recommended actions, and that this information can be used to choose less risky actions in unknown situations (Sect.~\ref{sec:results_unseen_situations}). Another potential use for the uncertainty estimation is to identify situations that should be added to the training process.
Further properties of the presented method are discussed in Sect.~\ref{sec:discussion}.

The main contributions of this paper are:
\begin{enumerate}
    \item{A novel application of an RL method for tactical decision-making in autonomous driving that can estimate the uncertainty of its decisions (Sect.~\ref{sec:experiments}).}
    \item{A criterion that determines if the trained agent is confident enough about a particular decision (\mbox{Sect.~\ref{sec:safeActions}}).}
    \item{A performance analysis of the introduced approach in different highway driving scenarios, compared to a commonly used heuristic method and a standard DQN approach (Sect.~\ref{sec:results}).}
\end{enumerate}



\section{Approach}
\label{sec:approach}

This section gives a brief introduction to RL, and a description of how the uncertainty of a recommended action can be estimated. The details on how this approach was applied to autonomous driving follows in Sect.~\ref{sec:experiments}.

\subsection{Reinforcement learning}

Reinforcement learning is a subfield of machine learning, where an agent interacts with some environment to learn a policy $\pi(s)$ that maximizes the future expected return~\cite{Sutton2018}. The policy defines which action $a$ to take in each state $s$. When an action is taken, the environment transitions to a new state $s'$ and the agent receives a reward $r$.
The reinforcement learning problem can be modeled as a Markov Decision Process (MDP), which is defined by the tuple $( \mathcal{S}, \mathcal{A}, T, R, \gamma)$, where $\mathcal{S}$ is the state space, $\mathcal{A}$ is the action space, $T$ is a state transition model, $R$ is a reward model, and $\gamma$ is a discount factor. At every time step $t$, the goal of the agent is to choose an action $a$ that maximizes the discounted return,
\begin{align}
    R_t = \sum_{k=0}^\infty \gamma^k r_{t+k}.
\end{align}

In Q-learning~\cite{Watkins1992}, the agent tries to learn the optimal action-value function $Q^*(s,a)$, which is defined as
\begin{align}
    Q^*(s,a) = \max_\pi \mathbb{E} \left[R_t | s_t = s, a_t = a, \pi\right].
\end{align}
The DQN algorithm uses a neural network with weights $\theta$ to approximate the optimal action-value function as $Q(s,a;\theta) \approx Q^*(s,a)$~\cite{Mnih2015}. Since the action-value function follows the Bellman equation, the weights can be optimized by minimizing the loss function
\begin{align}
    L(\theta) = \mathbb{E}_M \Big[ (r + \gamma \max_{a'} Q(s',a';\theta^-)
    - Q(s,a;\theta) )^2 \Big].
    \label{eq:loss}
\end{align}
The loss is calculated for a mini-batch $M$, and a target network $\theta^-$ is updated regularly.

\begin{algorithm}[!t]
    \caption{Ensemble RPF training process}\label{alg:training}
    \begin{algorithmic}[1]
        \For{$k \gets 1$ to $K$}
            \State Initialize $\theta_k$ and $\hat{\theta}_k$ randomly
            \State $m_k \gets \{\}$
        \EndFor
        \State $i \gets 0$
        \While{networks not converged}
            \State $s_i \gets $ initial random state
            \State $k \sim \mathcal{U}\{1,K\}$
            \While{episode not finished}
                \State $a_i \gets \argmax_{a} Q_k(s_i,a)$
                \State $s_{i+1}, r_i \gets $ \Call{StepEnvironment}{$s_i, a_i$}
                \For{$k \gets 1$ to $K$}
                   \If{$p \sim \mathcal{U}(0,1) < p_\mathrm{add}$}
                      \State $m_k \gets m_k \cup \{(s_i, a_i, r_i, s_{i+1})\}$
                   \EndIf
                   \State $M \gets $ sample from $m_k$
                   \State update $\theta_k$ with SGD and loss $L(\theta_k)$ 
                \EndFor
                \State $i \gets i + 1$
            \EndWhile
        \EndWhile
    \end{algorithmic}
\end{algorithm}

\begin{figure*}[!t]
    \centering
        \includegraphics[width=1.99\columnwidth]{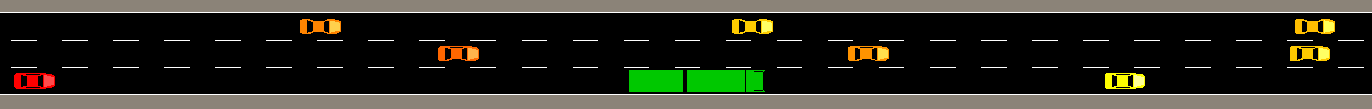}
        \caption{Example of an initial state of the highway driving scenario. The ego vehicle is shown in green, whereas the color of the surrounding vehicles represent their current speed. Yellow corresponds to $15$ m/s, red to $35$ m/s, and the different shades of orange represent speeds in between.} 
    \label{fig:trainingScenario}
\end{figure*}

\subsection{Bayesian reinforcement learning}   

The DQN algorithm returns a maximum likelihood estimate of the $Q$-values, but gives no information about the uncertainty of the estimation. The risk of an action could be represented as the variance of the return when taking that action~\cite{Garcia2015}. One line of RL research focuses on obtaining an estimate of the uncertainty by statistical bootstrap~\cite{Efron1982},~\cite{Osband2016}. An ensemble of models is then trained on different subsets of the available data and the distribution that is given by the ensemble is used to approximate the uncertainty. A better Bayesian posterior is obtained if a randomized prior function (RPF) is added to each ensemble member~\cite{Osband2018}. Then, each individual ensemble member, here indexed by $k$, estimates the $Q$-values as the sum
\begin{align}
    Q_k(s,a) = f(s,a;\theta_k) + \beta p(s,a;\hat{\theta}_k),
\end{align}
where $f$ and $p$ are neural networks, with parameters $\theta_k$ that can be trained, and parameters $\hat{\theta}_k$ that are kept fixed. The factor $\beta$ balances the importance of the prior function. When adding the prior, the loss function of Eq.~\ref{eq:loss} is changed to
\begin{align}
    \label{eq:loss_boot}
    L(\theta_k) = \mathbb{E}_M \Big[ & (r + \gamma \max_{a'} (f_{\theta^-_k}+\beta p_{\hat{\theta}_k})(s',a') \nonumber \\
    & - (f_{\theta_k}+ \beta p_{\hat{\theta}_k})(s,a) )^2 \Big].
\end{align}   

The full ensemble RPF method, which was used in this study, is outlined in Algorithm~\ref{alg:training}. An ensemble of $K$ trainable neural networks and $K$ fixed prior networks are first initialized randomly. A replay memory is divided into $K$ parallel buffers $m_k$, for the individual ensemble members (although in practice, this can be implemented in a memory efficient way that uses negligible more memory than a single replay memory). 
To handle exploration, a random ensemble member is chosen for each training episode. Actions are then taken by greedily maximizing the $Q$-value of the chosen ensemble member, which corresponds to a form of approximate Thompson sampling. 
The new experience $(s_i, a_i, r_i, s_{i+1})$ is then added to each ensemble buffer with probability $p_\mathrm{add}$. Finally, a minibatch $M$ of experiences is sampled from each ensemble buffer and the trainable network parameters of the corresponding ensemble member are updated by stochastic gradient descent (SGD), using the loss function in Eq.~\ref{eq:loss_boot}.

\subsection{Uncertainty threshold}
\label{sec:safeActions}


The coefficient of variation\footnote{The coefficient of variation is also known as the relative standard deviations, which is defined as the ratio of the standard deviation to the mean.} $c_\mathrm{v}(s,a)$ of the $Q$-values that are given by the neural networks can be used to estimate the agent's uncertainty of taking different actions from a given state. In this study, a hard uncertainty threshold $c_\mathrm{v}^\mathrm{safe}$ is used to classify if the agent is confident enough of its decision, but a progressive scale could also be used, which is further discussed in Sect.~\ref{sec:discussion}. When $c_\mathrm{v}(s,a) > c_\mathrm{v}^\mathrm{safe}$, action $a$ is considered unsafe in state $s$, which indicates that $(s,a)$ is far from the training distribution. The value of the parameter $c_\mathrm{v}^\mathrm{safe}$ can be determined by observing the performance of the agent and the variation in $c_\mathrm{v}(s,a)$ for the chosen action during testing episodes within the training distribution, see Sect~\ref{sec:results_training_scenarios} for further details.

When the training process is completed and the trained agent is deployed (i.e., not during the training process), the agent chooses actions by maximizing the mean $Q$-value of the $K$ neural networks, under the condition $c_\mathrm{v}(s,a) < c_\mathrm{v}^\mathrm{safe}$, i.e.,
\begin{equation}
    \begin{aligned}
        \argmax_{a} \frac{1}{K} \sum_{k=1}^K Q_k(s,a),\\
        \textrm{s.t.} \quad c_\mathrm{v}(s,a) < c_\mathrm{v}^\mathrm{safe}.
    \end{aligned}
\end{equation}
If no action fulfills the criteria, a fallback action $a_\mathrm{safe}$ is used.




\section{Implementation}
\label{sec:experiments}


A one-way highway driving scenario (Fig.~\ref{fig:trainingScenario}) was used to train and test the presented ensemble RPF algorithm. This section describes how the scenarios were set up, how the decision-making problem was formulated as an MDP, how the neural networks were designed, and how the training process was set up. The code that was used to implement the algorithm and experiments is available on GitHub \cite{sourceCode}.

\subsection{Simulation setup}
\label{sec:simulationSetup}


The Simulation of Urban Mobility (SUMO) traffic simulator was used for the experiments in this study~\cite{SUMO2018}. A one-way highway with three lanes was simulated, where the controlled vehicle consisted of a $16$ m long truck-trailer combination, with a maximum speed of $v_\mathrm{max}^\mathrm{ego}=25$ m/s. In the beginning of each episode, $25$ passenger cars were inserted into the simulation, with a random desired speed in the range $[v_\mathrm{min},v_\mathrm{max}] = [15, 35]$ m/s. In order to create interesting traffic situations, slower vehicles were positioned in front of the ego vehicle, and faster vehicles were placed behind the ego vehicle. Each episode was terminated after $N=100$ time steps, or earlier if a collision occurred or the ego vehicle drove off the road. The simulation times step was set to $\Delta t=1$ s.
Fig.~\ref{fig:trainingScenario} gives an example of the initial state of an episode.

The passenger vehicles were controlled by the standard SUMO driver model, which consists of an adaptive cruise controller for the longitudinal motion~\cite{Krauss1997}, and a lane change model that makes tactical decisions to overtake slower vehicles~\cite{Erdmann2015}. In the scenarios considered here, no strategical decisions were necessary, so the strategical part of the lane changing model was turned off. Furthermore, in order to make the traffic situations more demanding, the cooperation level of the lane changing model was set to $0$. Overtaking was allowed both on the left and right side of another vehicle, and each lane change took $t_\mathrm{lc}=4$ s to complete.


\subsection{MDP formulation\footnote{Technically, the problem is a Partially Observable Markov Decision Process (POMDP)~\cite{Kaelbling1998}, since the ego vehicle cannot observe the internal state of the driver models of the surrounding vehicles. However, the POMDP can be approximated as an MDP with a $k$-Markov approximation, where the state consists of the last $k$ observations~\cite{Mnih2015}. For this study, it proved sufficient to use only the last observation.}}

The decision-making problem was formulated according to the following Markov decision process.

\subsubsection{State space, $\mathcal{S}$}
The state of the system,
\begin{align}
    s = (\{x_i, y_i, v_{x,i}, v_{y,i}\}_{i\in0,\dots,N_\mathrm{veh}}),
\end{align}
consists of the positions $(x_i, y_i)$ and speeds $(v_{x,i}, v_{y,i})$ of each vehicle in a traffic scene, where index $0$ refers to the ego vehicle. The agent that controls the ego vehicle can observe the state of all surrounding vehicles within the distance $x_\mathrm{sensor}=200$ m.

\subsubsection{Action space, $\mathcal{A}$}
At every time step, the agent can choose between any combination of three longitudinal actions and three lateral actions, which consist of setting the acceleration to $\{\minus1, 0, 1\}$ m/s\textsuperscript{2} and $\{$\textit{`stay in lane', `change left', `change right'}$\}$. The final possible action is to brake hard by setting the longitudinal acceleration to $\minus4$ m/s\textsuperscript{2}. In total, this results in $10$ different possible actions.
Once a lane change is initiated, it cannot be aborted.
The fallback action $a_\mathrm{safe}$ is set to \textit{`stay in lane'} laterally and $\minus 4$ m/s longitudinally.

\subsubsection{Reward model, $R$}
The objective of the agent is to navigate through traffic in a safe and time efficient way. A simple reward model is used to achieve this goal. At every time step, the agent receives a positive reward of $1 - \frac{v_\mathrm{max} - v_0}{v_\mathrm{max}}$, which encourages a time efficient policy that, for example, overtakes slow vehicles. However, if a collision occurs, or the ego vehicle drives off the road (by changing lanes outside of the road), a negative reward of $r_\mathrm{col}=\minus10$ is added and the episode is terminated. Furthermore, if the behavior of the ego vehicle causes another vehicle to emergency brake, defined by a deceleration with a magnitude greater than $a_\mathrm{e}=\minus4.5$~m/s\textsuperscript{2}, or if the ego vehicle drives closer to another vehicle than a minimum time gap of $t_\mathrm{gap}=2.5$~s, a negative reward of $r_\mathrm{near}=\minus10$ is added, but the episode is not terminated. Furthermore, in order to discourage unnecessary lane changes, a negative reward of $r_\mathrm{lc}=\minus1$ is added when a lane change is initiated.

\subsubsection{State transition model, $T$} 
The state transition probabilities are implicitly defined by the generative simulation model, and not known to the agent.

\subsection{Neural network architecture}

A neural network estimates the $Q$-values of the different actions in a given state. The state $s$ is transformed to the normalized state vector $\xi$ before it is passed to the network, where all elements $\xi_*\in[\minus1,1]$. The positions and speeds of the surrounding vehicles are expressed as relative to the ego vehicle. Further details on how $\xi$ is calculated are given in Table~\ref{tab:neuralNetworkInput}.

\begin{table}[!t]
	\renewcommand{\arraystretch}{1.1}
	\caption{Input $\xi$ to the neural network.
	}
	\label{tab:neuralNetworkInput}
	\centering
	\begin{tabular}{ll}
		\toprule
	    Ego lane & $\xi_1 = 2y_0 / y_\mathrm{max} - 1$\\
	    Ego vehicle speed & $\xi_2 = 2v_{x,0} / v_\mathrm{max}^\mathrm{ego} - 1$\\
	    Lane change state & $\xi_3 = \sgn{(v_{y,0})}$\\
		Relative long. position of vehicle $i$ & $\xi_{4i+1} = (x_i-x_0) / x_\mathrm{sensor}$\\
		Relative lat. position of vehicle $i$ & $\xi_{4i+2} = (y_i-y_0) / y_\mathrm{max}$\\
		Relative speed of vehicle $i$ & $\xi_{4i+3} = \frac{v_{x,i} - v_{x,0}}{v_\mathrm{max} - v_\mathrm{min}}$\\ 
		Lane change state of vehicle $i$ & $\xi_{4i+4} = \sgn{(v_{y,i})}$\\
		\bottomrule
	\end{tabular}
\end{table}

In a previous paper~\cite{Hoel2018}, we introduced a one-dimensional convolutional neural network (CNN) architecture, which makes the training faster and gives better results than a standard fully connected architecture. 
By applying CNN layers and a max pooling layer to the part of the input that describes the surrounding vehicles, the output of the network becomes independent of the ordering of the surrounding vehicles in the input vector, and the architecture allows a varying input vector size.  

\begin{figure}[!t]
    \centering
        \includegraphics[width=0.8\columnwidth]{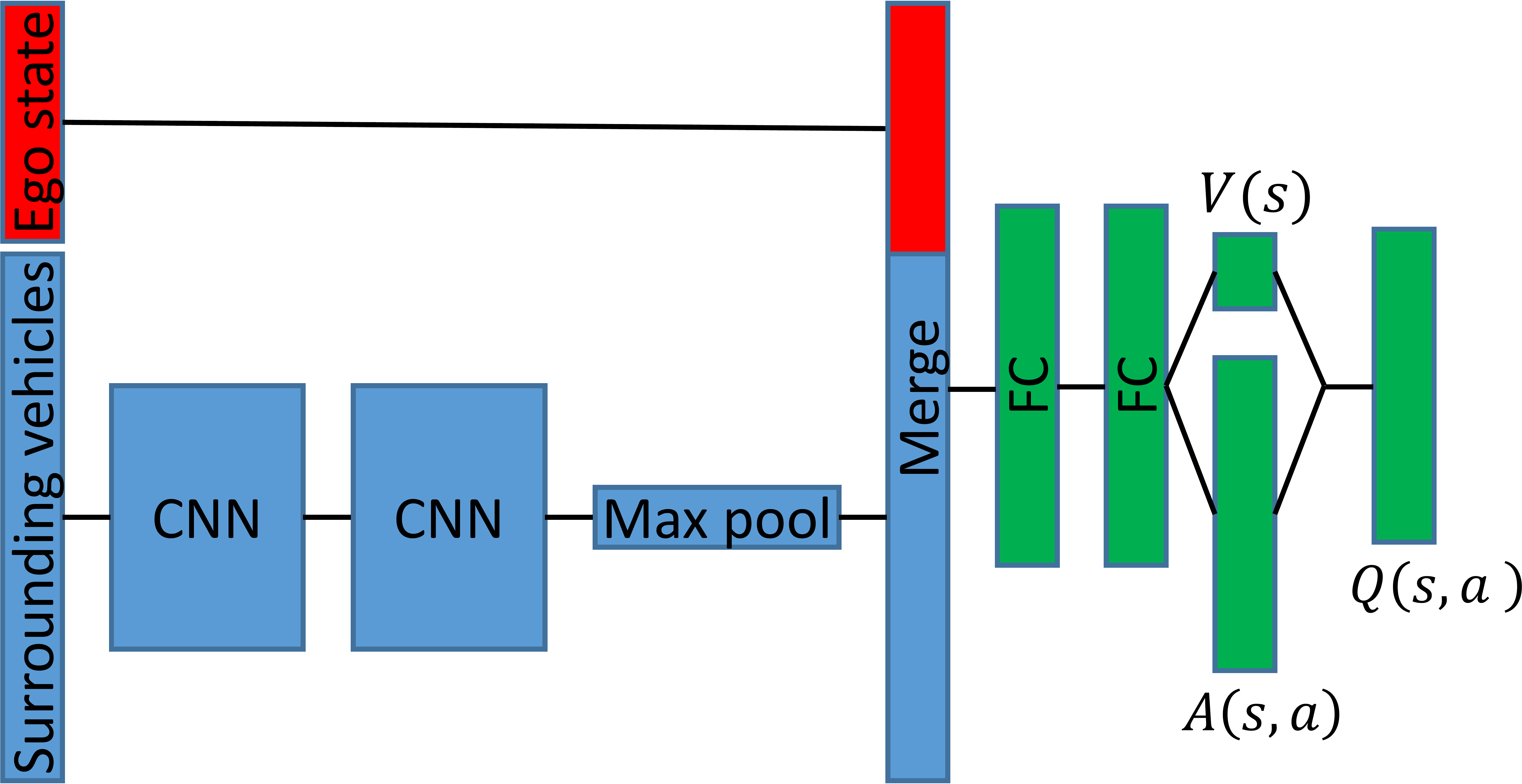}
        \caption{The neural network architecture that was used in this study.}
    \label{fig:neuralNetworkArchitecture}
\end{figure}

The neural network architecture that was used in this study is shown in Fig.~\ref{fig:neuralNetworkArchitecture}. 
Both convolutional layers have $32$ filters. The size and stride of the first convolutional layer is set to four, which equals the number of state inputs of each surrounding vehicle, whereas the size and stride of the second layer is set to one. The two fully connected (FC) layers have $64$ units each.
Rectified linear units (ReLUs) are used as activation functions for all layers, except the last, which has a linear activation function. The architecture also includes a dueling structure that separates the state value and action advantage estimation~\cite{Wang2016}.

\subsection{Training process}

Algorithm~\ref{alg:training} was used to train the ensemble of neural networks, with the exception that the loss function of Double DQN was used, which slightly changes the maximization term of Eq.~\ref{eq:loss} to $\gamma Q(s',\argmax_{a'} Q(s',a';\theta_i);\theta_i^-)$~\cite{Hasselt2016}. Adam~\cite{Kingma2014} was used to update the parameters $\theta_k$ of the $K$ ensemble members, and the update step was parallelized in order to speed up the process.
For episodes without collisions, the last experience was not added to the replay memory, in order to trick the agent that the episodes continue forever \cite{Hoel2018}.
Table~\ref{tab:hyperparameters} displays the hyperparameters of Algorithm~\ref{alg:training} and the training process. Due to the computational complexity, the hyperparameter values were selected from an informal search and not a systematic grid search.

\begin{table}[!bt]
	\renewcommand{\arraystretch}{1.2}
	\caption{Hyperparameters of Algorithm~\ref{alg:training} and baseline DQN.}
	\label{tab:hyperparameters}
	\centering
	\begin{tabular}{lr}
		\toprule
		Number of ensemble members, $K$ & $10$\\
		Prior scale factor, $\beta$ & $50$\\
		Experience adding probability, $p_\mathrm{add}$ & $0.5$\\
		Discount factor, $\gamma$ & $0.99$\\
		Learning start iteration, $N_\mathrm{start}$ & $50{,}000$\\
		Replay memory size, $M_\mathrm{replay}$ & $500{,}000$\\
		Learning rate, $\eta$ & $0.0005$\\
		Mini-batch size, $M_\mathrm{mini}$ & $32$\\
		Target network update frequency, $N_\mathrm{update}$ & $20{,}000$\\
		Huber loss threshold, $\delta$ & $10$\\

		\midrule

		Initial exploration constant, $\epsilon_\mathrm{start}$  & $1$\\
		Final exploration constant, $\epsilon_\mathrm{end}$ & $0.05$\\
		Final exploration iteration, $N_{\epsilon\mathrm{{\text -}end}}$ & $1{,}000{,}000$\\

		\bottomrule
	\end{tabular}
\end{table}

\subsection{Baseline methods}

The Double DQN algorithm was used as a baseline method (henceforth simply referred to as the DQN method). To make a fair comparison, the same hyperparameters as for Algorithm~\ref{alg:training} were used, with additional hyperparameters for an annealed exploration schedule, given in Table~\ref{tab:hyperparameters}.
During the testing episodes, the action with the highest $Q$-value was greedily chosen.
The standard SUMO driver model, which is further described in Sect.~\ref{sec:simulationSetup}, was used as a second baseline method.


\section{Results}
\label{sec:results}

The results show that the ensemble RPF method outperforms the SUMO driver model and performs more consistently than the baseline DQN method when tested on similar scenarios as the agents were trained on. However, when the trained agents were tested on scenarios outside of the training distributions, the ensemble RPF method both indicates a high uncertainty and chooses safe actions, whereas the DQN agent causes collisions. This section presents more details on the results, together with a brief analysis and discussion on some of the characteristics of the results, whereas a more general discussion follows in Sect.~\ref{sec:discussion}.

Both the ensemble RPF and the DQN agents were trained in a simulated environment (Sect.~\ref{sec:experiments}). At every $50{,}000$ added training sample, henceforth called training step, the agents were evaluated on $100$ different test episodes. These test episodes were randomly generated in the same way as the training episodes, but not present during the training. The test episodes were also kept identical for all the test phases and agents. The safety criterion described in Sect~\ref{sec:safeActions} was not active in the test episodes (since the uncertainty $c_\mathrm{v}$ varies during the training process), but used when the fully trained agent was exposed to unseen scenarios, see Sect.~\ref{sec:results_unseen_situations}.

\subsection{Within training distribution}
\label{sec:results_training_scenarios}



\begin{figure}[!t]
    \centering
        \includegraphics[width=0.98\columnwidth]{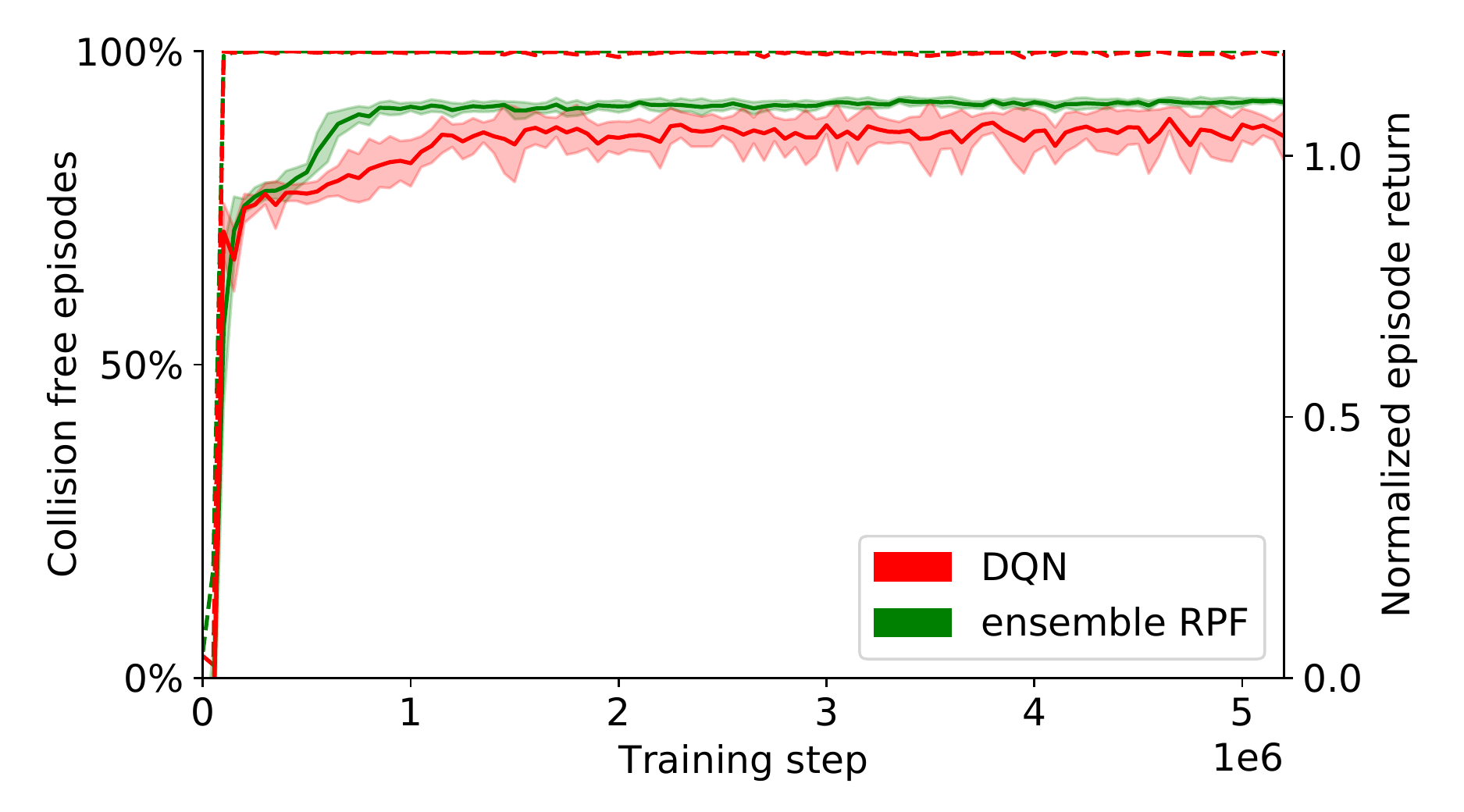}
        \caption{Proportion of collision free test episodes (dashed), and mean normalized return over training steps for the ensemble RPF and DQN agents (solid). The shaded areas show the standard deviation for $10$ random seeds.}
    \label{fig:rewardEvolution}
    \vspace{-12pt} 
\end{figure}

Fig.~\ref{fig:rewardEvolution} shows the proportion of collision free test episodes as a function of training steps for the ensemble RPF and DQN agents. For $10$ random seeds, the figure also shows the mean and the standard deviation of the return during the test episodes, normalized by the return of the SUMO driver. Both the ensemble RPF and DQN agents quickly learn to brake and thereby solve all test episodes without collisions, although the DQN agent experience occasional collisions later on during the training process. With more training, both methods also learn to overtake slow vehicles and outperform the SUMO driver model. The ensemble RPF agent receives a slightly higher return and has a more stable performance compared to the DQN agent. The small variation in final performance between random seeds of the ensemble RPF agent is likely due to that a close to optimal policy has been found. \looseness=-1

\begin{figure}[!t]
    \centering
        \includegraphics[width=0.98\columnwidth]{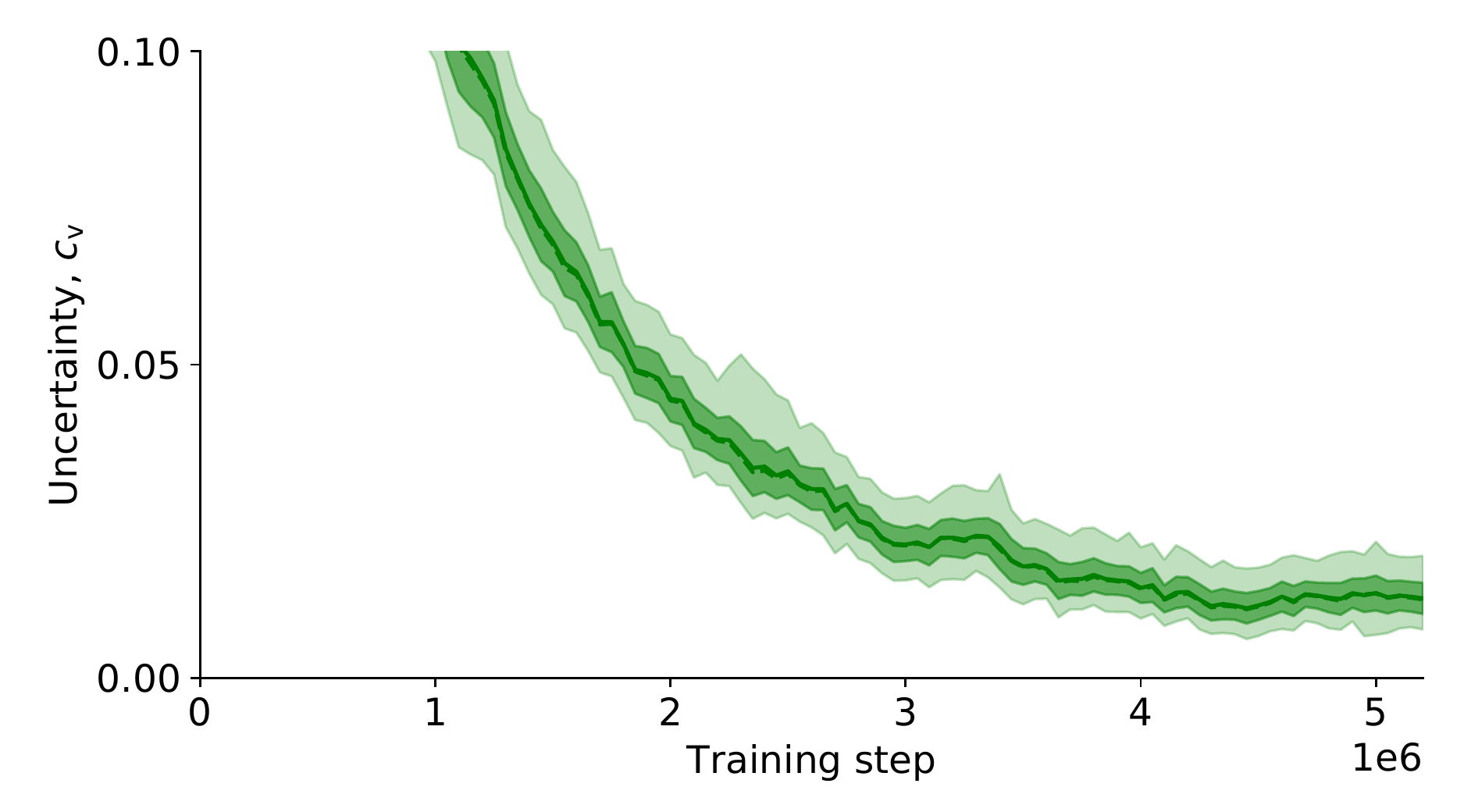}
        \caption{Mean uncertainty, represented by the coefficient of variation $c_\mathrm{v}$, of the chosen action during the test episodes. The dark shaded area represent the standard deviation and the bright shaded area represent percentiles $1$ to $99$.}
    \label{fig:uncertaintyEvolution}
    \vspace{-8pt} 
\end{figure}

To gain insight in the uncertainty estimation during the training process, and to illustrate how to set the uncertainty threshold parameter $c_\mathrm{v}^\mathrm{safe}$ (Sect.~\ref{sec:safeActions}), Fig.~\ref{fig:uncertaintyEvolution} shows the coefficient of variation $c_\mathrm{v}$ for the chosen action during the test episodes as a function of training steps. Note that the figure shows the uncertainty of the chosen action, whereas the uncertainty for other actions could be higher. After around four million training steps, the coefficient of variation settles at around $0.01$, with a small spread in values, which motivates setting $c_\mathrm{v}^\mathrm{safe}=0.02$.

\begin{figure*}[!t]
    \centering
        
        \begin{subfigure}[]{1.99\columnwidth}
            \includegraphics[width=0.99\columnwidth]{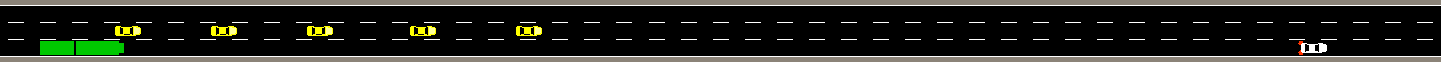}%
            \vspace{-1pt}
            \includegraphics[width=0.99\columnwidth]{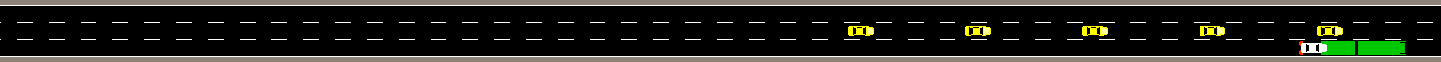}%
            \vspace{-1pt}
            \includegraphics[width=0.99\columnwidth]{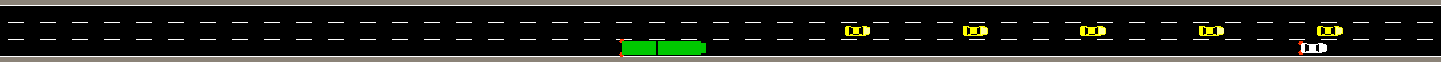}
            \vspace{-2pt}
            \caption{Accident situation with a stopped vehicle, shown in white.}
        \label{fig:unseen_accident}
        \end{subfigure}
        
        \vspace{5pt}
    
        \begin{subfigure}[]{1.99\columnwidth}
            \includegraphics[width=0.99\columnwidth]{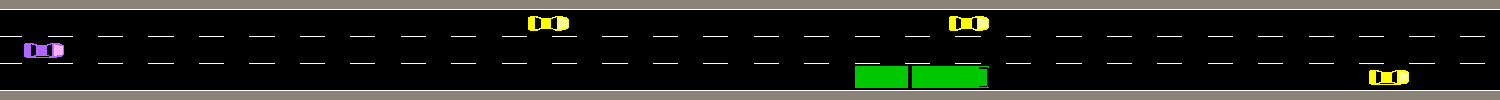}%
            \vspace{-2pt}
            \includegraphics[width=0.99\columnwidth]{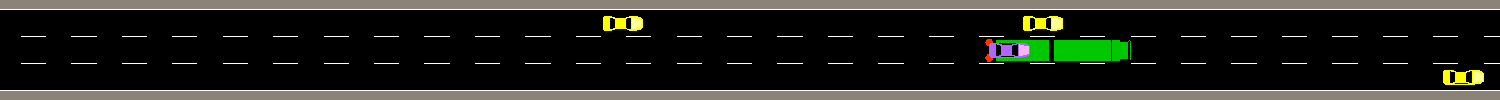}
            \vspace{-2pt}
            \includegraphics[width=0.99\columnwidth]{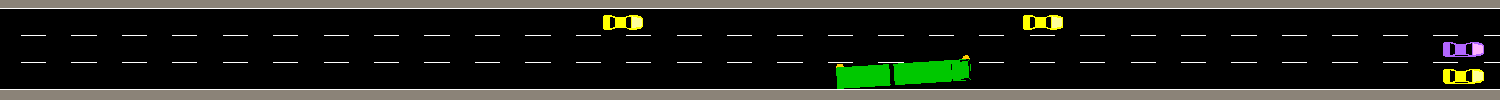}
            \vspace{-2pt}
            \caption{Situation with a speeding vehicle, shown in purple.}
        \label{fig:unseen_speeding}
        \end{subfigure}
        
        \caption{Two situations that are outside of the training distribution and cause collisions if the confidence of the agent is not considered. The top panel of each figure shows the initial state, whereas the two bottom panels show the state for the DQN and ensemble RPF agents, after $12$ s in (a) and $7$ s in (b).}
    \label{fig:unseenSituations}
    \vspace{-15pt} 
\end{figure*}

\subsection{Outside training distribution}
\label{sec:results_unseen_situations}



In order to illustrate the ability of the ensemble RPF agent to detect unseen situations, the agent that was obtained after five million training steps was deployed in scenarios that were not included in the training episodes. In various situations that involve an oncoming vehicle, the uncertainty estimate was consistently high, $c_\mathrm{v}>0.2$. This level is one order of magnitude larger than the uncertainty threshold $c_\mathrm{v}^\mathrm{safe}$, and therefore clearly indicates that such situations are outside of the training distribution.

\begin{figure}[!t]
    \centering
        \includegraphics[width=0.99\columnwidth]{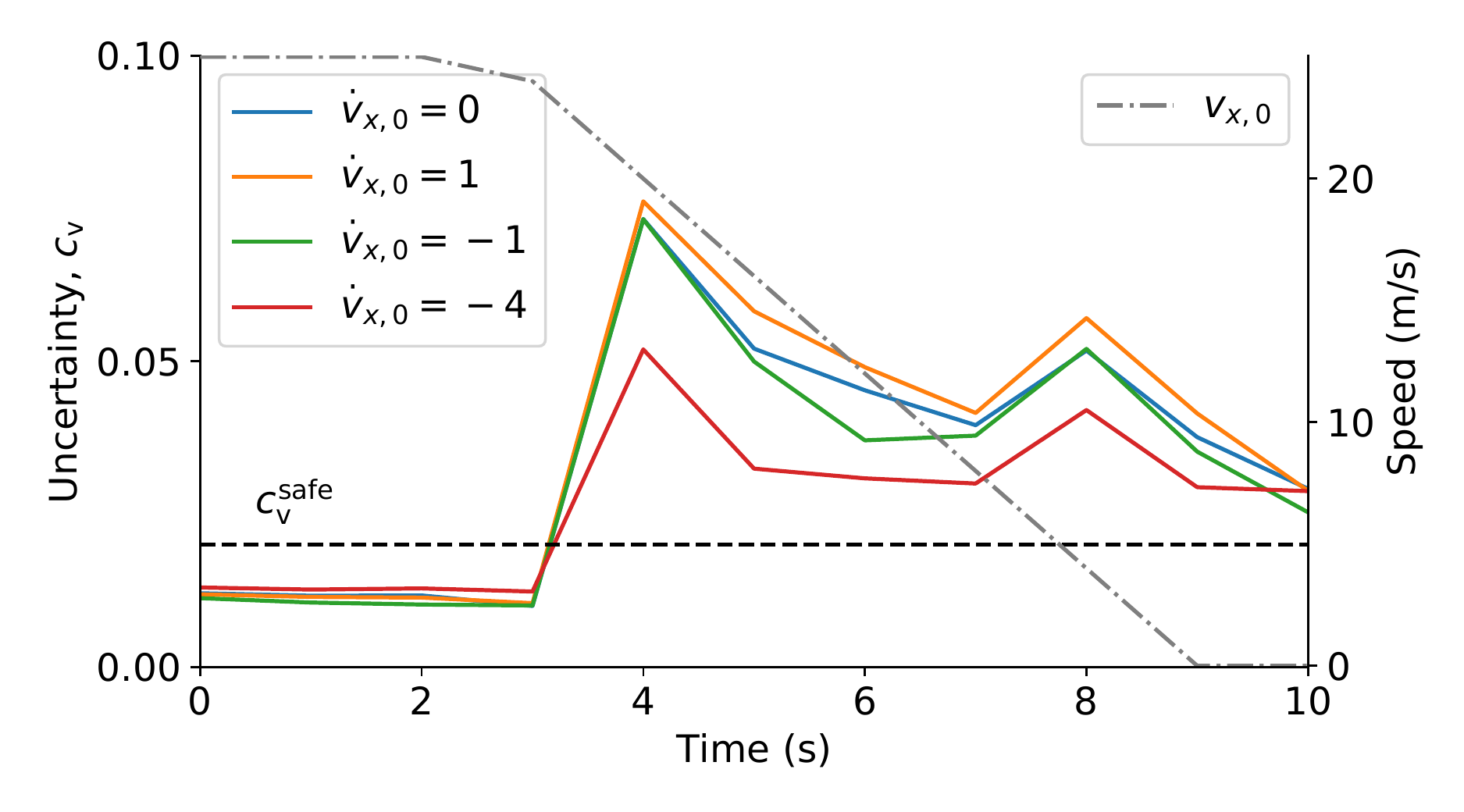}
        \caption{Uncertainty $c_\mathrm{v}$ for the \textit{'stay in lane'} action with different accelerations, and ego vehicle speed $v_{x,0}$, during the case with a stopped vehicle. The uncertainty for the other possible actions are orders of magnitude larger. Due to the sensor range limitation, the stopped vehicle is not observed until $t=   4$ s.}
    \label{fig:cv_standstill}
    \vspace{-5pt}
\end{figure}

For a deployed agent that has been trained in one-way highway traffic, an arguably more representative situation that the agent could be exposed to involves an accident, which has caused a vehicle to stop on the highway, see Fig.~\ref{fig:unseen_accident}.
As mentioned in Sect.~\ref{sec:simulationSetup}, the surrounding vehicles were simulated with a random speed in the range $[15,35]$ m/s during the training, hence a vehicle that stands still is outside of the training distribution. The ego vehicle starts with a speed of $25$ m/s and is placed in the rightmost lane, with the stopped vehicle $300$ m in front of it. There are several slower vehicles in the center lane, which makes changing lanes impossible. The DQN agent does brake when it approaches the stopped vehicle, but since such a situation was not present in the training episodes, the agent does not brake early enough to avoid a collision. The ensemble RPF agent also outputs the highest $Q$-value for maintaining its current speed when the ego vehicle is far from the stopped vehicle, and would have collided if that action had been chosen. However, as soon as the stopped vehicle is within the ego vehicle's sensor range $x_\mathrm{sensor}$, the uncertainty $c_\mathrm{v} > c_\mathrm{v}^\mathrm{safe}$, see Fig.~\ref{fig:cv_standstill}. Since this uncertainty level indicates that the situation is outside of the training distributions, the agent chooses to brake hard (Sect.~\ref{sec:safeActions}) early enough to avoid a collision. \looseness=-1

The trained agent was also tested in a situation that involves a speeding vehicle, see Fig.~\ref{fig:unseen_speeding}, where a vehicle that drives at $55$ m/s is approaching the ego vehicle from behind, in the neighboring lane. A slow vehicle is positioned in front of the ego vehicle, which creates an incentive to overtake it on the left. The DQN agent does change lanes to the left, but then causes a collisions since the speeding vehicle cannot brake fast enough. The ensemble RPF agent also estimates the highest $Q$-values for the lane changing actions, but estimates $c_\mathrm{v} > 0.025$ for changing lanes, which is larger than the uncertainty threshold $c_\mathrm{v}^\mathrm{safe}$, until the speeding vehicle has passed. However, $c_v \approx 0.015$ for staying in the lane and braking with $\minus 1$ m/s\textsuperscript{2}, which the agent then does, according to the policy that was described in Sect.~\ref{sec:safeActions}. When the speeding vehicle has passed, the ego vehicle changes lanes, since $c_\mathrm{v}$ has then decreased to below the threshold.

Videos of the presented scenarios and additional situations, together with the code that was used to obtain the results, are available on GitHub \cite{sourceCode}.




\section{Discussion}
\label{sec:discussion}








The results show that the ensemble RPF algorithm can be used to train an agent that is aware of the uncertainty of its decisions. The ensemble RPF agents outperforms both the DQN agent and the heuristic SUMO driver model within the training distribution (Fig.~\ref{fig:rewardEvolution}). In addition, the ensemble RPF agent can also indicate its uncertainty when the agent is exposed to situations that are far from the training distribution (Fig.~\ref{fig:unseenSituations}). 
In this study, the uncertainty information was used to choose safe actions, by prohibiting actions with a level of uncertainty that exceeds a defined threshold. However, to formally guarantee functional safety with a learning-based method is challenging and likely requires an underlying safety layer in a real application \cite{Underwood2016}. While the presented approach could reduce the activation frequency of such a safety layer, a possibly even more important application could be to improve the learning process.
The uncertainty information could be used to guide the training to situations that the agent is currently not confident about, which could improve the sample efficiency and broaden the distribution of situations that the agent can handle.
Furthermore, if an agent is trained in a simulated environment and later deployed in real traffic, the uncertainty information could be used to detect situations that need to be added to the simulated world.

Since a simple safety function was used in this study, a hard uncertainty threshold $c_\mathrm{v}^\mathrm{safe}$ was used to determine if the agent is confident enough to take a particular action. If a more advanced safety function would receive the information from the agent, it could be beneficial to instead output a continuous confidence measure. One option is to define such a confidence measure as $1 - \frac{c_\mathrm{v}(s,a) - c_\mathrm{v}^\mathrm{min}}{c_\mathrm{v}^\mathrm{safe} - c_\mathrm{v}^\mathrm{min}}$, where $c_\mathrm{v}^\mathrm{min}$ is a parameter that defines the minimum uncertainty. A value of $1$ would then indicate maximum confidence, and values below $0$ would be considered unsafe.

The main disadvantage of using the ensemble RPF method compared to DQN is the higher computational cost, since $K$ neural networks need to be trained instead of one. However, the design of the algorithm allows an efficient parallelization of the training process, which in practice reduces the effect. Both agents were trained on a desktop computer, where the DQN agent required $36$ hours and the ensemble RPF agent required $72$ hours to complete five million training steps. Osband et al. reports that the difference can be further reduced to $20\%$ in wall-time with their implementation~\cite{Osband2016}.



\section{Conclusion}
\label{sec:conclusion}


The advantage of using Bayesian RL compared to standard RL for tactical decision-making in autonomous driving has been demonstrated in this paper. The ensemble RPF method learns to make more efficient decisions and it has a more stable performance compared to the DQN method within the training distribution. Outside of the training distribution, the ensemble RPF method is aware of the high uncertainty and can fall back to taking safe actions, in order to avoid collisions. However, since the DQN agent does not possess the uncertainty information, collisions occur in unknown situations.

A possibly even more important aspect when having information on what the agent knows and does not know is that the training can be adapted accordingly. To investigate this further is a topic for future work. The performance of the proposed method will also be further evaluated in different types of traffic situations in a future paper.

\balance








\linespread{0.93}
\bibliographystyle{IEEEtran}
\bibliography{IEEEabrv,references}

\end{document}